# Keynote ESWEEK 2017: Small Neural Nets Are Beautiful: Enabling Embedded Systems with Small Deep-Neural-Network Architectures

Forrest Iandola and Kurt Keutzer
DeepScale and
UC Berkeley EECS
forrest@deepscale.ai, keutzer@berkeley.edu

## ABSTRACT

Over the last five years Deep Neural Nets have offered more accurate solutions to many problems in speech recognition, and computer vision, and these solutions have surpassed a threshold of acceptability for many applications. As a result, Deep Neural Networks have supplanted other approaches to solving problems in these areas, and enabled many new applications. While the design of Deep Neural Nets is still something of an art form, in our work we have found basic principles of design space exploration used to develop embedded microprocessor architectures to be highly applicable to the design of Deep Neural Net architectures. In particular, we have used these design principles to create a novel Deep Neural Net called SqueezeNet that requires as little as 480KB of storage for its model parameters. We have further integrated all these experiences to develop something of a playbook for creating small Deep Neural Nets for embedded systems.

## 1 INTRODUCTION

Over the last 50 years, in the diverse areas of natural language processing, speech recognition, and computer vision, progress has been achieved through the orchestration of dozens of diverse algorithms generally classified under the heading of "machine learning." In just the last five years the best results on most of the problems in these areas have been provided by a single general approach: Deep Neural Networks (DNNs). Moreover, for many problems, such as object classification and object detection, results using DNNs enabled computer vision algorithms to offer an acceptable level of accuracy for the first time. Thus, in many application areas, broader algorithmic exploration is being replaced by the creation of a single DNN architecture[1].

Compared to other software architectures, DNNs are quite simple. They consist of a simple feedforward pipe-and-filter software architecture whose computation is statically-schedulable and whose control is input-independent. The precise execution of the DNN depends heavily on a set of static constants called *weights* or *model parameters*. Nevertheless, the particular organization of the DNN and the precise characterization of the computations in the filter elements is diverse enough to create a rich design space. Thus, in the creation of a DNN to solve a particular application problem there are two implicit questions:

1) What is the right DNN architecture? In other words, what characteristics are we looking for in our DNN architecture?

2) How do we find the right DNN architecture?

For much of the computer vision community, "the right" DNN architecture has been the DNN that produced the highest accuracy results on a target standard benchmark set, such as ImageNet [8]. As for the means to find that architecture, strategies to find "the right" architecture have principally remained in the minds in the most experienced DNN designers. Much of the leading-edge work on DNN design has occurred within major corporations that use Deep Learning, and while some of these companies have been generous in revealing the details of their architectures, the deeper design intuitions, as well as the full disclosure of how to successfully train their DNNs to fully replicate their results, has not always been provided. Thus, for many wishing to apply DNNs to their application, the norm has become to simply take pre-trained DNN architectures developed by others and, if necessary, augment the training of these DNNs with their own application-specific datasets.

Our work has countered these directions in a couple ways. First, for us "the right" DNN is one that offers acceptable accuracy but operates in real-time within power and energy constraints of its target embedded application. This focus has led us away from experimenting with DNN architectures with a large (e.g. 30M or more) number of model parameters as their memory footprint makes them prohibitively expensive for deployment in embedded systems. Instead, we have chosen to explore the other extreme: very small DNN architectures capable of fitting in even the smallest embedded systems.

---

[1] We use the term "DNN architecture" precisely because we're exploiting the analogy with computer architectures. Do not confuse our DNN architectures with the architectures of hardware accelerators for computing DNNs.

Second, we also approached the question "what is the right DNN architecture?" from an embedded systems perspective. A systematic approach to design-space exploration of microprocessor architectures for embedded applications must surely be one of the high-water marks of research in embedded systems. Thus, as we approached the second question we sought to leverage decades of research on systematic design-space exploration of application-specific embedded microprocessors.

The first significant result of our efforts on design space exploration of DNNs for embedded systems was SqueezeNet [25]. SqueezeNet is a DNN targeted for the object classification problem. With just 1.2 million model parameters that can be stored in 4.8 MB, SqueezeNet achieved the same accuracy as the popular DNN AlexNet that initially sparked the recent explosion of interest in DNNs. AlexNet uses 60M model parameters that require 240MB for storage. Thus, SqueezeNet provided a 50x reduction in the number of model parameters compared to AlexNet. We were able to further reduce the model size by using Deep Compression techniques [14] for pruning and quantization. With these techniques, we reduced the model size of SqueezeNet to a surprising 480KB.

Space does not permit a self-contained tutorial on DNN architectures and to fully understand what follows we suggest the reader consult [10] for general concepts of DNNs and review notes for the Stanford course entitled *Convolutional Neural Networks for Visual Recognition* (CS231) for basic concepts of Deep/Convolutional Neural Networks.

## 2 WHAT IS THE RIGHT DNN ARCHITECTURE?

### 2.1 How bigger became better

As is common in academia, progress in many areas of machine learning has been marked relative to a publicly available set of standard benchmarks. In the particular case of computer vision, for many researchers in the field recent progress has been measured by accuracy on image classification as measured with regard to the ImageNet benchmark set [8]. The ILSVRC-2010 ImageNet challenge consists of a training set of 1,200,000 labeled images with 1000 object categories. The validation or test set on which accuracy is measured consists of 200,000 photographs. Generally speaking, progress on image classification had been slow but steady for many years. Measured on the ILSVRC-2010 benchmark, for example, accuracy improvements from 2010 to 2011 were approximately 2.5%. In 2012, Krizhevsky introduced the AlexNet DNN architecture that resulted in a 10% improvement over the prior year's best efforts and demonstrated about the same percentage improvement compared to AlexNet's runner up in the 2012 competition [28]. With eight layers AlexNet was not much deeper than prior neural nets, but AlexNet had 60 million model parameters, requiring 240MB of storage. With the success of AlexNet, a growing field of computer vision and machine learning researchers became focused on using successively more intricate Deep Neural Nets to solve image classification and other problems.

While the efforts of the large computer vision community are diverse, for many, the answer to the question "What is the right DNN architecture?" is "the architecture that produces the best accuracy on a topical benchmark set." Moreover, for most researchers, accuracy is not only the preeminent concern, it is the exclusive concern, and the energy required to compute the DNN is rarely, if ever, reported. As a result, a principal trend was toward larger DNNs that provided the best accuracy.

### 2.2 Why Small is Beautiful

In his book entitled *Small is Beautiful* [42], E. F. Schumacher proposes to counter the dominant economic trend that "bigger is better," in steel mills for example, with smaller coordinated industrial efforts. We brought this inspiration to our work on developing DNNs. In particular, we sought to counter the trend toward larger general-purpose DNN architectures by exploring DNN architectures that would be better suited to particular embedded applications. As we began to provide DNNs that were indeed useful in embedded applications, users of our DNNs came forward with other advantages of small DNNs that we detail below. As we shall see, over the entire life cycle of a DNN, from training to embedded deployment, small networks provide a number of key advantages over larger networks.

*2.2.1 Small DNNs are more feasible for embedded implementation.* For some Internet-of-Things (IOT) applications, as well as many other embedded applications, the 95MB storage requirements of even a relatively small DNN architecture such as Inception V3 [47] are prohibitively expensive. Mid-end FPGA devices such as the Altera Max 10 have only ~6 MB of user accessible on-chip memory. Thus, many embedded applications absolutely require small DNNs with very few (e.g 2M) model parameters and modest (e.g. 6MB) overall storage requirements.

*2.2.2 Freedom from the cloud: Low latency, privacy, and "always on" reliability.* The power/energy trade-offs between computing an application in the cloud and computing it exclusively on the client are complex; however, in many embedded applications there are other advantages to running application on the client. First, if an application can run locally with efficiency then most often overall latency will be reduced. Second, however well-grounded are their concerns, many customers do not want photos or speech recordings sent up to the cloud. Thus, to alleviate privacy concerns client-centric processing for speech recognition or photo processing applications may be desirable. Finally, most of us have experienced the frustration of using mobile applications that work well when our cellular handset is connected to WiFi but fail entirely as soon as WiFi is lost. Client-centric processing of applications can supply always-on reliability, or at least graceful degradation when WiFi is not available.

*2.2.3 Making over-the-air (OTA) updates more feasible.* Most mobile applications benefit from "in field" updates of the software. When DNNs are used, "the software" becomes the model parameters of the DNN. In automated driving, companies such as Tesla periodically copy new models from their servers to customers' cars. Generally, this process is called an "over-the-air (OTA) update." With AlexNet, this would require 240MB of communication from the server to the car. On the other hand, smaller models require less communication, making frequent updates more economical. Industry experts assert that cellular communications comprise 60-70% of the cost of popular vehicle telematics programs [3]. As more and more functions are provided by DNNs then reductions in DNN size translate directly to reductions in the communication costs of OTA updates.

*2.2.4 Smaller DNNs train faster.* To train a DNN on a large dataset (e.g. ImageNet [8]) in a few hours (rather than days or weeks) can require using distributed computing. In our own experience with highly-distributed DNN training [24], the problem quickly becomes communication bound, and the communication bottleneck is the distributed communication of the gradient updates associated with the DNN model parameters. Reductions in the number of DNN model parameters give corresponding reductions in the total training time.

*2.2.5 Smaller DNNs can be faster and more energy efficient.* Since Krizhevsky's work, the dominant computing platform for computer vision has been to use one or more GPUs each with a Thermal Design Power (TDP) in excess of 200 Watts. In contrast, mobile handsets must operate in TDPs under 10 Watts and make a minimal draw on handset battery reserves. Thus, DNNs that are power/energy efficient are required for these mobile applications.

On the one hand computations of DNNs are statically schedulable and the speed and energy requirements for running a DNN on an application workload should be easily analyzable. On the other hand, the precise speed and energy of an application is tightly interleaved with application constraints, such as fixed latency requirements, and target processor platform characteristics, such as on-chip memory sizes. In many embedded applications that we have consider, the batch size is naturally 1. For example, in style transfer, discussed later in this paper, a user snaps a single picture and transfers to that single picture a style from a set of alternatives. In embedded applications that we have considered with real-time constraints, such as object detection from a video camera, latency requirements dictate an input batch size of 1 to meet the application requirements of processing 30 video frames per second.

As for target platform constraints: in the computing platforms we considered the energy cost of an off-chip memory access is 100x that of an arithmetic computation [18]. Suffice it to say, for the purposes of this paper, smaller models have fewer off-chip memory accesses resulting in increases in speed and reductions in energy. These advantages come to the forefront when the cost of memory accesses to fetch model parameters cannot be amortized over larger input batch sizes. As a particular example Han shows how reductions in model parameters due to Deep Compression *alone* produce speedups of 3-4x and energy reductions of 3-7x [14]. Further reductions due to smaller models can produce larger reductions.

Moreover, whatever the advantages of small DNNs are today, we believe that the large investment in DNN accelerators will only increase these advantages. These accelerators promise to dramatically reduce the costs of DNN computations, which will in turn only make the costs of memory accesses relatively higher. Also, we anticipate that future DNN accelerators will have sufficient on-chip memory to fully accommodate both the model parameters of smaller DNNs and the results of their intermediate activations.

## 3 HOW DO WE FIND THE RIGHT DNN ARCHITECTURE? CHALLENGES IN THE CURRENT APPROACH TO DNN DESIGN

### 3.1 Benchmarks and metrics

For the DNNs that we discuss here, a benchmark set consists of a *training data set* that is used to set the precise values to the model parameters of the DNN during training, and an additional *validation data set* that is used to evaluate the results of running the DNN. The recent progress in the application of DNNs to computer vision has been widely attributed to 1) more sophisticated DNN architectures, 2) greater computing power for both training and inference phases, and 3) large data sets for training DNNs. Of these three factors, the training data set is the most commonly overlooked, and the sizes and diversity of training datasets has not evolved as quickly as other aspects of the research field [7]. Private companies guard their training data, and the gap between public and private benchmark sets creates an ever-bigger gap between the "haves" and the "have nots."

That academic researchers would rather create benchmarks for new problems rather than evolve existing benchmark sets is not surprising. Mainstream research in areas such as computer vision are strongly motivated by the desire to fundamentally advance the project of creative visions systems that match or surpass human comprehension of their environment. Practically speaking, at any one time this broad objective is reduced to progress on a sub-problem such as object classification, and this progress is measured by accuracy on a particular benchmark set, such as ImageNet [8]. When an acceptable level of accuracy is achieved on such a problem area, then rather than evolving the benchmark set associated with that problem, the mainstream of the field moves onto another, harder problem, such as automating the creation of captions in images. Often the inauguration of a new research direction is coincidental with the creation a new set of benchmarks. Thus, for most researchers, *accuracy* on the benchmarks of one of the current "hot topics" is the primary metric.

As for metrics: a recent paper by Huang *et al.* notes "only a small subset of papers discuss running time in any detail" [22]. While a growing number of researchers are beginning to consider the speed of DNN computations, the metrics of speed and energy efficiency that are necessary for practical realization of computer vision advances are largely ignored by mainstream computer vision researchers. Some have opted to use a static calculation of the number of computations performed by a DNN as a proxy for speed; such metrics fail to consider the impact of memory accesses on the actual execution of the computation.

## 3.2 Defining, evaluating, and exploring the design space of DNNs

Each DNN architecture forms one point in the design space of DNN architectures. Frameworks such as Caffe [27], Caffe2, MXNET [7], or Tensorflow [1] have made it easy to define a particular DNN architecture.

Properly evaluating a single design point, that is, a DNN architecture, entails training the DNN architecture on a training data set, and then evaluating the accuracy, speed, and energy of the model on a validation data set. The frameworks above also facilitate the training of DNNs and their implementation. Nevertheless, there are a number of significant challenges entailed in evaluating a DNN architectures. The first is appropriately setting the hyperparameters used during training, the second is scaling the training to multiprocessors to reduce training time, and the third is understanding fine-tuning techniques to improve accuracy during inference.

If the design of an individual DNN is an art, it is during the training and inference phases that Deep Learning earns the title of a "black art." Hyperparameters to be considered during training include topics such as strategies for initializing the model-parameter weights and setting the learning rate. In computer vision applications there are many techniques, such as data augmentation, used to improve the utility of the training set of images. Further, there are other techniques used to increase the accuracy during the inference stage. These include multi-crop, horizontal flipping, and others [21].

To compound all these challenges, given a single DNN architecture, and a benchmark set the size of ImageNet, it can take hours to weeks on a single CPU/GPU to perform the evaluation of just one design point. Thus, to rapidly evaluate design points requires access to a computing cluster, and the expertise to efficiently use it in the distributed training of a DNN.

For all these reasons, the dimensionality of design space explored by current researchers and practitioners varies widely. For leading edge researchers, the design space is only limited by their own creativity and progress is made through a mixture of the introduction of new DNN architectural ideas, such as the Inception module [46], and the revival of older ideas such as Long Short-Term Memory (LSTM) [17].

However, for many mainstream practitioners using DNNs, the state-of-the-art is very different. Many DNN architectures are made open source with their model parameters available but missing the full details on the hyperparameters necessary for training the DNN from scratch to achieve the highest accuracy on a data set. In response to this, even relatively sophisticated practitioners opt to simply use a predefined DNN and adapt it to their application through additional training (known by the respectable term of "transfer learning"), or by making minimal changes to the final layers.

## 4 FINDING A DISCIPLINED APPROACH TO DESIGN-SPACE EXPLORATION

After surveying the state of the art in DNN design in 2015, we aimed to explore some alternative directions motivated by our experience with embedded systems. First, with regard to the question "What is the right DNN architecture?" we saw a lot of value in putting DNN-based computer vision applications on mobile embedded systems such as cellular handsets, but our initial estimates indicated to us that even the more economical of the popular DNN architectures were too large and too slow to compute on these devices. Thus, we first set out to find smaller DNN architectures suitable for embedded applications, and then investigated the extent that these smaller DNNs could still retain the accuracy of larger models.

Second, to counter some of the challenges of the design of DNN architectures, we aimed to apply what we had learned about taking a disciplined approach to designing application-specific instruction processor (ASIP) architectures for embedded applications.

In the past, the design of processors and microprocessors was also an undisciplined art form. In fact, as late as 2002, a survey of the architectures of over thirty network processors showed them to lack unifying architectural principles [43], even though most were focused on a common application: packet forwarding. Nevertheless, the architectures of some microprocessors and application-specific instruction processors for embedded systems have reflected the use of best practices in design-space exploration. These practices were comprehensively surveyed by Gries [11], and the underlying principles and best-practices were mined from these examples and presented in the MESCAL methodology [12]. Iandola's dissertation [23] sought to generalize this MESCAL methodology of design space exploration to the design of Deep Neural Nets. The first fruit of the application of this methodology was SqueezeNet [25], which will be discussed in more detail below. Rephrased for the design of DNN architectures, the MESCAL methodology became:

1. Judiciously defining benchmarks and metrics to evaluate DNNs
2. Inclusively defining the DNN design space
3. Efficiently defining and evaluating points in the DNN design space
4. Systematically exploring the design space of DNN architectures

In the following we present the best practices we've gathered for performing design-space exploration of DNN architectures.

### 4.1 Judiciously defining benchmarks and metrics

In the last five years, DNNs have enabled computer vision researchers to within a level of acceptability for an ever-growing set of practical applications for which computer vision represents a key enabling technology. These include automated driving, security and surveillance, virtual reality, as well as a variety of social-media related applications. To be truly useful for these applications now entails trading off a number of metrics:

- Accuracy
- Model size
- Energy and power efficiency
- Inference speed (i.e. deployed runtime)
- Training latency

Our only nuance is to especially emphasize the importance of choosing benchmarks that are well tailored to the specifics of our target application, and then holistically measuring the results according to the metrics named.

## 4.2 Inclusively defining the DNN Design space

In our approach, we place a high-value on inclusively defining the precise design space of DNN architectural elements that we wish to explore, and then systematically exploring that space. The architectural elements include the number of layers in the DNN, the type and dimensions of each layer, and the organization of layers. We will discuss in more detail the aspects of the DNN architectural space we explore in Section 5.

## 4.3 Efficiently Defining and Evaluating Design Points

The first element of our approach to evaluating design points has been to simply describe as clearly as possible the precise procedure for evaluating a design point, such as hyperparameter choices, so that results can be replicated. A second element of our approach has been to devote significant effort to developing the FireCaffe [24] system that allowed us to use a sizable cluster of GPUs to speed the evaluation of each design point. GPUs and library support for DNN computations have gotten much faster since we initially embarked on this work, and we find that what used to take an hour on 128 NVIDIA K20 GPUs now takes an hour on just 8 NVIDIA Titan X-Pascal GPUs. Moreover, each of the popular DNN frameworks is putting an emphasis on automating distributed training. These developments should provide encouragement for more researchers and practitioners to experiment with design-space exploration.

## 4.4 Systematically Exploring the Design Space

In the systematic exploration of a design space for a DNN the first step is to create a representative benchmark set. The next step is to identify clear goals with respect to the metrics: accuracy, model size, and other metrics such as speed. A meaningful set of goals for the design of a DNN might be:

- model parameters and intermediate activations must fit into a 8192KB SRAM
- the DNN model must achieve 20% top-5 error rate, or less, on the problem of object classification
- a frame rate of 15 FPS must be achieved and a frame rate of 30 FPS is highly desirable
- energy used should be less than 2 Joules/frame on the targeted computing platform

Our next task is to identify a DNN model, large or not, that gives the desired accuracy. Following that, the designer must identify the parts of the initial DNN that are over-provisioned (*e.g.* have more model parameters than needed). We refer to the point where adding more parameters to the model (or to a layer of the model) does not lead to an accuracy improvement as a "saturation point." The better we understand the saturation points in the design space, the easier it is to identify ideal dimensions of neural network layers. To search for the saturation points, we train several versions of a neural network such that each network has a unique tradeoff in the design space.

A full list of the techniques that we apply to reduce size is given in Section 5 below. For example, as we will see in Section 5.3, when we replace some of the 3x3 filters with 1x1 filters, we find that we can replace half the 3x3 filters with 1x1 filters before accuracy starts to degrade, and this is a saturation point.

To better understand how this general procedure is translated into a more fine-grained grid search, refer to Section 5.1 of [25].

## 4.5 Our Design Space Exploration Discovers SqueezeNet

As we mentioned in Section 1, the focus on DNNs for computer vision was sparked in large part by a DNN winning the ImageNet image classification competition by a large margin in 2012. This model, AlexNet [28], achieves 80.3% top-5 accuracy on the 1000-category ImageNet image classification dataset with 60 million (240MB) model parameters. AlexNet and its successors were too large and required too much computation to be useful for embedded applications such as object detection on cellular handsets. As AlexNet was already used as a standard reference for comparison, we posed ourselves our own answer to Question 1: The right DNN is one that is amenable to implementation in embedded systems such as cellular handsets and at least matches AlexNet in accuracy.

Our search for such a model followed the basic principles of this section. As a benchmark, we used the ImageNet classification benchmark. Initially our only metric was model size. At each perturbation of DNN architecture we applied transformations that we describe in Section 5 below. To rapidly evaluate DNN we used our own FireCaffe [24] system. Altogether we scaled training and evaluation to 128 GPUs per experiment (with multiple 128-GPU experiments running concurrently on an 18000-GPU supercomputer) to evaluate each design point and explored over 1000 different DNN design points.

Our first interesting result was SqueezeNet [25]. Fulfilling our initial design objectives, SqueezeNet achieved the same accuracy as AlexNet with just 1.2 million model parameters requiring 4.8 MB of storage. This amounted to a nearly 50x reduction in the number of parameters compared to AlexNet. SqueezeNet achieves this result by applying a series of strategies including limiting the number of 3x3 filters in the network ("kernel reduction"), limiting the total number of filters in the network ("channel reduction"), and applying the remaining filters in each layer to more input data (achieved through "evenly-spaced downsampling"). We describe both the techniques used to achieve this and the details of the resulting architecture in more detail in Section 5 below, in Iandola's dissertation [23] and in the SqueezeNet paper [25].

Reducing the number of parameters by 50x relative to AlexNet while retaining the same accuracy was our first achievement with SqueezeNet. We were able to further reduce the memory requirements by collaborating with Song Han, and using the Deep Compression techniques he developed [14]. With these techniques, we reduced the model size to a surprising 480KB, which is 500x smaller than uncompressed AlexNet.

## 4.6 SqueezeNet: The Phenomenon and its Limitations

We were happy that we realized our own design objectives in SqueezeNet, and we were especially pleased with the modest number of model parameters in the final post-Deep Compression version. That said, we were quite aware of the rapid progress in DNN design, and we felt that the design-space exploration approach used to discover SqueezeNet, and the orchestration of size reduction techniques, would be a more important contribution than the DNN itself.

Nevertheless, soon after its creation SqueezeNet found its way into a variety of DNN design frameworks, including Caffe, TensorFlow, and most recently Caffe2; SqueezeNet is even provided as a "default" reference DNN architecture in some of these frameworks. Confirmation of SqueezeNet's amenability to implementation on embedded processors was soon demonstrated by its implementation on the Qualcomm's Snapdragon 820 [37] and ARM processors [45]. SqueezeNet was also demonstrated on NXP processors in the company's demo booths on the tradeshow floor of the Embedded Vision Summit.

SqueezeNet has also served as an inspiration to other DNN designs, and, without exaggeration, a new DNN that acknowledges itself as a relative of SqueezeNet appears every week. We will cite a couple particularly interesting and representative design studies. Gschwend [13] does a very nice design study that uses SqueezeNet as the basis for his own DNN topology, ZynqNet. Gschwend is able to pack both his DNN and its activations onto the on-chip SRAM of the FPGA. If not the most sophisticated use of SqueezeNet, at least the most amusing, is its role in creating Version 2 of the popular application "Not Hotdog" seen on the HBO TV Series Silicon Valley. In fact, Tim Anglade's description of the creation of that app is itself a truly interesting design study [19].

As a DNN, SqueezeNet has two principal deficiencies. The first is that its classification accuracy lags the rapidly advancing state-of-the art in accuracy. This is indeed an important limitation; however, for many applications, such as style transfer, this has little impact. The problem of *style transfer* is defined as taking an image from a camera and putting it into the style of a particular painting such as The Starry Night by Vincent van Gogh. Style transfer is popular with users on photo sharing and social networking platforms. One of the early implementations of style transfer was based on the VGG-19 DNN architecture [9] and a public implementation was provided by Johnson [35]. More recently Johnson's code has been modified by Zeng to use SqueezeNet instead of VGG [31]. SqueezeNet has 120x fewer parameters than VGG-19, and whatever the loss due to "accuracy" in this case, the amenability of the SqueezeNet-based model to implementation on mobile handsets more than compensates for it.

While SqueezeNet dramatically reduces the number of model parameters relative to AlexNet, it does not significantly reduce the total number of computations. While memory accesses are more expensive than computations, computations are not free, and memory spills during computations may lead to additional memory accesses. In short, a second deficiency of SqueezeNet is the relatively large number of computations. An approach to addressing this problem in small nets has been demonstrated by later work such as MobileNets [20], described in more detail below.

## 5 A PLAYBOOK FOR CREATING SMALL DNNs

Our design-space exploration discovered SqueezeNet; however, progress in DNN design has proceeded rapidly, and one and a half years later we have a better perspective on what are the key elements of creating a small DNN. We summarize these in this section.

## 5.1 Background on DNN dimensions

We begin with a review of some of the key concepts of DNNs, but this is meant to be a refresher, not a stand-alone tutorial.

*5.1.1. Layers* Deep neural networks are comprised of a sequence of layers. A layer is simply a function that is applied to input data. Some layers have predefined functions such as interpolation (e.g. pooling) or mechanisms for zeroing out negative numbers (e.g. ReLU *[34]*). In other layers such as convolutional layers, a filter is applied to the input of that layer and the model parameters of the filter are learned during training. To produce the final probabilities for a problem such as image classification, early networks such as AlexNet *[28]* used fully-connected layers. A fully-connected layer is a special type of convolution layer where the output is a vector instead of a 3D tensor. In object classification, the fully-connected layer has as many filters as object categories to be classified (e.g. 1000).

*5.1.2. Filter dimensions* A convolutional DNN (or CNN) has $n$ layers, numbered *i=0:(n-1)*, built almost entirely out of convolutional and ReLU layers. Each convolutional layer has multiple filters. In a typical case, the filters in a convolutional layer have a spatial resolution of 3x3. Each filter also has multiple channels, so a filter can simultaneously "see" information across multiple channels and multiple pixels. A key point is that the number of filters in layer *i* defines the number of input channels in layer *(i+1)*. The dimensions of filters are hyperparameters that are typically selected by an experimenter. The dimensions of filters have a direct impact on a DNN's accuracy, model size, energy efficiency, and so on. With this in mind, filter dimensions pose a useful avenue for design space exploration with the goal of identifying ideal DNN models for embedded systems.

*5.1.3. Activations and Downsampling* Activations are the temporary variables passed from one DNN layer to the next. In convolutional DNNs, activations are 3-dimensional, where the dimensions are width, height, and channels. Convolutional networks for image classification typically downsample the activations every few layers, which has the effect of taking (for example) a 640x480x3 image as input and producing a 1x1xC vector as the output of the last layer, where C is the number of categories (*e.g.* the ImageNet-1k dataset has 1000 categories, so C=1000). This downsampling is sometimes called "pooling." In today's convolutional models, it is common to configure the pooling layers to halve the height and width of their input activations. Finally, note that activations are temporary variables that are unique to each image, and activations do not contribute to the quantity of model parameters. With these basic definitions, we now proceed to review the key elements of creating small DNNs.

## 5.2 Replacing Fully-connected Layers with Convolution Layers

Fully-connected (FC) layers are a special case of convolution where the output data is not a tensor but rather a 1D vector, *i.e.* 1x1x(#Channels) instead of HxWx(#Channels). In popular DNNs such as AlexNet [28] and VGG [44], the majority of the parameters are in the fully-connected layers. The mere presence of fully-connected layers is not the culprit for the high model size, rather, the problem is that some FC layers of VGG and AlexNet have a huge number of channels and filters. For example, the FC7 layer in AlexNet has 4096 input channels and 4096 filters; this alone results in 67MB of floating-point parameters.

In SqueezeNet, we found that we could delete FC layers that have many filters, replacing these with convolution layers with fewer filters, without affecting the accuracy. Just the FC7 layer of AlexNet has 14 times more parameters than an entire SqueezeNet model.

## 5.3 Kernel Reduction

In number of popular DNN models such as VGG and AlexNet, almost every convolutional layer is configured such that the filters have a spatial resolution of at least 3x3. However, the DNN architecture Network in Network [32] showed that 1x1 filters (technically, 1x1x(#Channels)) are also a useful building block for DNNs. While 1x1 filters cannot see outside of a 1-pixel radius, they retain the ability to combine and reorganize information *across* channels.

For reasons mentioned earlier, we are interested in small models, and a 1x1 filter has 9x fewer parameters than a 3x3 filter. This led us to the following question: If we replace some of the 3x3 filters in a layer with 1x1 filters, we know it will reduce the model size, but how will this change affect accuracy? To evaluate this question, we devised a layer template that has a predefined number of filters, some of which are 1x1 and some of which are 3x3. This layer template is a simplification of the Inception module [46] that we call a *Fire Module* [25]. We further defined a metaparameter variable $p$, which denotes the percentage of filters in the layer which are 3x3. We use this layer template several times in the same model, and we use the same value of $p$ across [all relevant layers in the model]. We generated versions of the model with different values of $p$, and we trained each version from scratch on ImageNet.

What we discovered from this experiment is that there is a saturation point in $p$. When $p$ = [50%, 67.5%, 75%, 82.5%, or 100%], the top-5 ImageNet error rate of the model is 15%. That is to say, we can double the number of 3x3 filters in the model (from $p$=50% to $p$=100%), without any measurable increase to accuracy. Given that most researchers do not report doing a search for the saturation point in their models, we imagine that many DNN architectures may be overparameterized, that is, they have extra parameters that only serve to increase the model's memory and storage footprint. The process of replacing high-resolution (*e.g.* 3x3) filters with low-resolution (e.g. 1x1) filters has come to be known as "kernel reduction."

## 5.4 Channel Reduction

In our design space exploration that led to the discovery of SqueezeNet, we found that, even after doing kernel reduction, the 3x3 filters still contributed heavily to the total quantity of parameters. Recall from Section 5.1, the number of filters in layer $i$ defines the number of input channels to layer $i+1$. So, in addition to limiting the number of 3x3 filters, to achieve small model size we also needed to limit the number of input channels to each 3x3 filter. We refer to decreasing the number input channels as "channel reduction." In other papers, this approach is sometimes called a "bottleneck layer."

## 5.5 Evenly-spaced Downsampling

Another design consideration in DNNs for computer vision is the dimensionality (particularly height and width) of the activations produced by the convolution layers. It is the placement of downsampling (*e.g.* pooling) layers in the network that affects the dimensions of the height and width of the activations. Putting many pooling layers at the beginning of the network leads to few activations for most layers in the network. Conversely, putting most of the pooling layers near the end of the network leads to many activations for most layers in the network.

Recall from Section 5.1.3 that the height and width of the activations produced by each neural network layer doesn't affect the number of model parameters, but the dimensions of the activations do affect the accuracy, and the computational complexity of the DNN.

To understand how activation size contributes to tradeoffs between accuracy and computational complexity, we generated a number of SqueezeNet-like models, where each model has a unique set of positions for the pooling layers. From this design space exploration, we observed that downsampling too early in the network hurts accuracy. We found that downsampling too late in the network didn't diminish accuracy, but it increased the network's computational complexity. In the context of SqueezeNet, we found the ideal choice was to space out the downsampling as evenly as possible over the convolutional layers.

## 5.6 Depthwise Separable Convolutions

We mentioned in Section 5.1 that the filters in convolution layers have an associated height and width (e.g. 3x3), as well as a number of channels. A common default setting is for the number of channels in each filter in layer $L$ to be equal the number of input channels $C$ in layer $L$. However, an innovation of DNNs such as the MobileNets family [20], is that an additional dimension *g* is defined. The number of channels per filter is defined as *C/g*.

If *g=2*, each filter is applied to *C/2* channels. Specifically, half the filters are applied to channels 0 to (*C*/2-1), and the other half of the filters are applied to (*C*/2 to *C*-1). If *g=C*, then each filter is applied to just one channel. Changing *g*=1 to *g*=2 leads to half the number of parameters in a filter. Changing *g*=1 to *g=C* leads to *C* times fewer parameters in a filter. In networks from the MobileNets family, *C* can be as high as 1024, so in certain layers setting *g=C* can lead to a 1024x reduction in the number of parameters.

There is not yet a standard name for convolutions with *g*>1, but this concept is sometimes called *depthwise separable convolutions* [20], *cardinality* [51], *filter groups* [26], or *group convolutions* [49].

## 5.7 Shuffle Operations

After replacing many of the 3x3 filters with 1x1 filters (*kernel reduction*), reducing the total number of filters (*channel reduction*), replacing fully-connected layers with convolutions, and factorizing 3x3 filters (*separable convolutions*), the result is a much smaller model. In our experience, most of the parameters in this new model are in the remaining 1x1 convolution filters. Very recently, researchers have begun to investigate how to factorize 1x1 filters to further reduce the quantity of parameters. One option would be to again do depthwise separable convolutions (*g*>1) for the 1x1 filters. However, if we have a sequence of 1x1 and 3x3 layers, and all of these layers have (*g*>1), we essentially have *g* independent networks that do not exchange data at all. To address this problem, a recent paper [53] proposed the following approach in a DNN called ShuffleNet: use (*g*>1) for 1x1 and 3x3 filters, but perform a shuffle permutation on the channels after some layers. This allows the 1x1 filters to be have a channel dimension of (*C/g*) instead of *C*, which reduces the parameter count, yet it avoids splitting the NN into a number of independent networks that do not exchange data. We see the factorization of 1x1 convolutions to further reduce parameter count as interesting area of future research.

## 5.8 Pruning, Compression, and Distillation

That connections between neurons in a neural net could be *pruned* was illustrated early on by LeCun [29]. This results in the elimination of the model parameters associated with the connection. Also, *quantization* and *coding techniques* to reduce computing and storage requirements have long been studied in Digital Signal Processing. Nevertheless, the integrated application of all these techniques in such a way as to reduce the storage requirement of a DNN *while retaining accuracy* is not so obvious, and this has been explored by Han *et al.* in their work on Deep Compression [14]. Deep Compression reduced the storage requirements of SqueezeNet by a factor of 10x and has shown even greater reductions on other, larger, DNNs [15]. A related approach to *model compression* is *distillation* [5] in which knowledge is transferred from a larger model to a smaller model. Quantization is taken to an extreme in work by Rastegari *et al.* where binary values for model parameters (Binary-weight-networks) and binary weights for both inputs and model parameters (XNOR-Net) are explored [38]; however, accuracy is impacted relative to full precision networks.

## 5.9 Playbook Reflections

In this section we have summarized the principle techniques for developing a small DNN. Removing fully connected layers, kernel reduction, channel reduction, evenly spacing downsampling, and Deep Compression are all techniques that we used in the design-space exploration that produced SqueezeNet. Depthwise separable convolutions, shuffle operations, and distillation were not integrated into our approach. The integrated orchestration of all these techniques into a single integrated design-space exploration procedure is a work in progress.

# 6 A BROADER TREND OF REDUCING THE SIZE OF DNN MODELS

Small model size is particularly well motivated in embedded systems; however, reducing the size of models reduces memory and, nearly always, computational requirements. In this section we briefly review other trends in model reduction.

For the ImageNet competition in 2014, more accurate models were developed. The VGG-19 [44] and GoogLeNet-v1 [47] models both achieved half the prior error rate (i.e. achieved top-5 error rate of 10%) relative to the best models reported in 2012. Further, while the VGG-19 model contains 575MB of parameters, the GoogLeNet-v1 model contains just 54MB of parameters. To achieve this, the GoogLeNet-v1 model used 1x1 convolution filters alongside higher-resolution filters, that is, the smaller model makes use of *kernel reduction*.

For the 2015 edition of the ImageNet competition, error rates of the winning models again halved, with the most accurate models achieving roughly 5% top-5 error. These most accurate model was ResNet-152 (241MB) [16]; the defining characteristics of this model are its depth (over 150 layers) and its "residual" shortcut connections that send data forward over multiple layers. However, the Inception-v3 model achieves equivalent accuracy using only 96MB of parameters [47]. The Inception-v3 uses kernel reduction, substituting some 3x3 filters with smaller 1x1 filters while retaining accuracy.

# 7 RELATED RESEARCH AND APPLICATIONS OF SMALL NEURAL NETWORKS

In the previous two sections we have described our procedure for exploring the design space for DNNs, and we have explained what we feel are the key architectural principles that lead to the creation of smaller DNNs. We used our own work, SqueezeNet, as a running example to convey our personal experiences applying these processes and principles. Meanwhile, a growing portion of the DNN research community has begun to develop DNN models that are smaller compared to mainstream DNN models. We briefly survey some of these developments below.

## 7.1 Small Models for Image Classification

Earlier in the paper, we talked about AlexNet, a 240MB DNN model which won the ImageNet 2012 competition by classifying images with a top-5 error rate of approximately twenty percent. Newer models have achieved an equivalent or slightly better error rate with fewer model parameters. These models include Network-in-Network (30MB) [32], TinyNet (7.9MB) [52], and SqueezeNet (4.8MB) [25]. Layers in the smaller models have fewer filters and/or fewer channels than the larger models, that is, the smaller models use *channel reduction* described above. Recently, models including "A Compact DNN" [49] and MobileNets [20] used *depthwise separable convolutions* to reduce model parameters while retaining accuracy. In "A Compact DNN," each 3x3 filter has ¼ the number of channels as the layer's input data, so each 3x3 filter is applied to ¼ of the input data, and this leads to a 16.4MB model with similar accuracy to GoogLeNet-v1 and VGG. The MobileNets authors report several versions of the model that they developed in the course of their design space exploration, and the most resource efficient models have just one channel in each 3x3 filter; the versions of the MobileNets models that have VGG-like accuracy have 10.4MB to 16.8MB of parameters. Beyond having a small model size, some versions of MobileNets also require 30x less computation than VGG-19 while producing similar accuracy.

## 7.2 Small Models for Object Detection

Modern object detection DNNs are broken into two parts: a "backbone" NN and a "detector" NN. The backbone is often borrowed from an image classification NN. The detector NN is designed specifically for the problem of object detection. In widely-cited works such as Faster R-CNN [40], YOLO [39], and SSD [33], the detector NN contains 100MB or more of parameters. In the ConvDet detector NN that we recently proposed in our SqueezeDet paper [48], fully-connected layers are replaced with convolution layers, and channel reduction is applied to these convolution layers, leading to as little as 3MB of parameters in the detector NN. In one implementation, the SqueezeDet model couples the SqueezeNet backbone with the ConvDet detector, and the whole model has just 7.9MB of parameters. Anisimov *et al.* recently took a similar approach of experimenting with mixing different backbone feature extractors NNs and detector NNs to arrive at a number of small object detectors [2]. Recently, Li used *distillation* to reduce DNN model sizes for object detection by a factor of 4 while retaining accuracy [30]. The MobileNets family also includes small DNNs for object detection.

## 7.3 Small Models for Other Problems in Computer Vision

Small models have propagated to other problems in computer vision. Here we will just mention a couple. On the computer vision problem of *semantic segmentation*, the popular SegNet [4] model has 117MB of model parameters, but the ENet [36] model achieves similar accuracy with just 0.7MB of parameters. The ENet network relies especially on *channel reduction*, so it has few filters and channels per layer.

Xiao *et al.* recently advanced the state-of-the-art accuracy on the problem of *Chinese character recognition* [50]. Further, while previous models had at least 23MB of parameters, the new model proposed by Xiao *et al.* has just 2.3MB of parameters. This work relies especially on model compression techniques to achieve this small model size.

## 7.4 Further Reading and Open Problems

*7.4.1 Further Reading* Various dimensions for comparing small and resource-constrained DNNs is discussed by Canziani [6].

*7.4.2 Open Problems* We have taken a very pragmatic orientation to the problem of building small models, but underlying our explorations are a number of fundamental questions: Are small nets more or less prone to overfitting? Just how small can a DNN get and still be useful? For example, how small can a DNN be and still get top-5 error rate of 3% on the ImageNet classification benchmark? A sharp answer to these questions may lie in the application of algorithmic information theory to neural nets. Such ruminations are not new (*Cf. [41]*), and progress on these questions is slow.

## 7.5 Commercial Applications

While we have motivated our work by embedded applications, we have primarily focused on published research. It is worth noting that a variety of commercial embedded computer-vision applications are appearing that can run locally on a mobile client. For example, the Facebook Camera app has a variety of effects that utilize style transfer and image segmentation and these effects appear to be able to run offline on the client. The mobile Google Translate app can run offline on the client. Perhaps in response to growing interest in client-centric apps, Clarifai has developed a Mobile SDK (software developer kit) that can run online or offline on mobile handsets. We suspect that all of the above are powered, at least in part, by small DNNs.

## 8 CONCLUSIONS

The mainstream computer vision community, like many subfields of AI and machine learning, will naturally continue to try to discover new methods to give computing machines human-like abilities to comprehend their environment. The work described in this paper takes another direction: adapting existing DNN-based applications of computer vision to practical use within the constraints of embedded applications.

In this paper we have aimed to clearly identify and articulate the numerous advantages of DNNs with fewer model parameters for embedded applications, particularly embedded computer vision applications. We have surveyed the current state of the art to identify many techniques that may be applied in the design of small DNNs, and we have described a general design-space exploration procedure that may be used to coordinate the application of these techniques. We have used our SqueezeNet DNN as an example throughout, but we have also tried to acknowledge the research efforts of a growing research community that shares the common goal of developing DNNs for embedded and mobile applications.

When we started this paper we hoped to comprehensively review the recent developments associated with small DNNs. Now, facing conference page limitations, we only hope to have given the reader some indication of why we think that small DNNs are, indeed, beautiful.

## ACKNOWLEDGMENTS

Thanks to Amir Gholaminenjad, Tom Meyer, Pete Warden, and especially Song Han for comments and input on the paper. This work was partially supported by the DARPA PERFECT program, Award HR0011-12-2-0016, together with ASPIRE Lab sponsor Intel, as well as lab affiliates HP, Huawei, Nvidia, and SK Hynix. This work has also been partially sponsored by BMW, Intel, and the Samsung Global Research Organization.